\documentclass[11pt]{article}
\usepackage{arxiv}

\usepackage{amsmath,amssymb,amsfonts}
\usepackage{graphicx}
\usepackage[hidelinks]{hyperref}
\usepackage[numbers,sort&compress]{natbib}
\usepackage{amsmath,amssymb,amsfonts}
\usepackage{graphicx}
\usepackage{hyperref}
\usepackage{hyperref}
\usepackage{url}

\usepackage{graphicx} 
\usepackage{amsmath,amssymb,amsthm,mathtools}
\usepackage{bm}
\usepackage{bbm}
\usepackage{enumitem}
\usepackage{subcaption}
\usepackage{float}

\newcommand{\R}{\mathbb{R}}
\newcommand{\N}{\mathbb{N}}

\newcommand{\e}{\mathrm{e}}

\newcommand{\dd}{\mathrm{d}}

\newcommand{\abs}[1]{\left\lvert #1\right\rvert}
\newcommand{\cL}{\mathcal{L}}

\theoremstyle{plain}
\newtheorem{theorem}{Theorem}
\newtheorem{proposition}[theorem]{Proposition}

\theoremstyle{definition}

\theoremstyle{remark}

\title{Learning Heat-Based Equations in Self-Similar Variables}

\author{Shihao Wang \\
Department of Mathematics\\
University of Wisconsin--Madison\\
Madison, WI 53706, USA \\
\texttt{swang2232@wisc.edu} \\
\And
Qipeng Qian\\
Program of Applied Mathematics, \\
Department of Mathematics, \\
University of Arizona \\
Tucson, AZ 85721, USA \\
\texttt{qqian@arizona.edu}
\And
Jingquan Wang \\
Department of Mechanical Engineering \\ 
University of Wisconsin-Madison\\
Madison, WI 53706, USA \\
\texttt{jwang2373@wisc.edu}
}

\begin{document}
\maketitle

\begin{abstract}
We study solution learning for heat-based equations in self-similar variables (SSV). We develop an SSV training framework compatible with standard neural-operator training. We instantiate this framework on the two-dimensional incompressible Navier-Stokes equations and the one-dimensional viscous Burgers equation, and perform controlled comparisons between models trained in physical coordinates and in the corresponding self-similar coordinates using two simple fully connected architectures (standard multilayer perceptrons and a factorized fully connected network). Across both systems and both architectures, SSV-trained networks consistently deliver substantially more accurate and stable extrapolation beyond the training window and better capture qualitative long-time trends. These results suggest that self-similar coordinates provide a mathematically motivated inductive bias for learning the long-time dynamics of heat-based equations.
\end{abstract}

\section{Introduction}

Neural-network-based methods for solving PDEs have advanced rapidly in recent years. Physics-informed neural networks (PINNs) \cite{raissi2019pinn} enforce governing equations through the loss, while operator-learning architectures such as DeepONets and Fourier neural operators learn nonlinear maps between function spaces and perform well on parametric benchmarks \cite{li2021fno,lu2021deeponet,takamoto2024pdebench,li2023pino}. However, for time-dependent problems, long-time extrapolation is often fragile: models trained on a finite horizon can accumulate error and suffer from distribution shift under rollout, leading to instability beyond the training window.

Existing approaches for improving long-horizon rollout stability include
(i) observation-informed extrapolation that injects PDE structure or sparse measurements into the prediction pipeline \cite{zhu2023reliable};
(ii) refinement/correction mechanisms that iteratively denoise, correct, or project predictions back to physically consistent states \cite{lippe2023pderefiner,huang2025physicscorrect,wei2025inc};
(iii) learnable time-integration and continuous-time operator formulations that couple operator learning with numerical time stepping or temporally consistent surrogates \cite{nayak2025tideeponet,abueidda2026time_resolution_independent,diab2025tno};
and (iv) rollout-aware training protocols (e.g., recurrent/self-rollout or curriculum schemes) that mitigate exposure bias by aligning training with inference-time dynamics \cite{ye2025rno,ghule2025dccl}.
In this paper, however, we take a representation-level approach guided by physical intuition: we exploit the intrinsic scaling structure of the governing dynamics and work in variables that better reflect its long-time behavior. This viewpoint reshapes the learning problem into a form aligned with the system’s underlying physical principles, and is complementary to the above techniques.

In particular, we focus on ``heat-based'' evolution equations, i.e., systems whose linear part is the heat equation. Prototypical examples include the two-dimensional incompressible Navier-Stokes equations and the one-dimensional viscous Burgers equation. For such systems, long-horizon prediction is especially challenging because the characteristic length scale grows while the amplitude decays, causing the effective scales of the solution to change substantially over time. Since most neural-operator methods are formulated in physical coordinates $(x,t)$, training on a fixed spatial box and a finite time horizon can capture only a diminishing fraction of the relevant dynamics as $t$ increases.

Self-similar variables (SSV) provide a natural remedy to this mismatch as classical theories \cite{MR1017289, giga1988large, kato1994,gallay_wayne_2002,kim_tzavaras_2001} show that the solutions admit a self-similar long-time behavior. Under the change of variables \(  \xi = x/{\sqrt{t+1}},\  \tau = \log(t+1) \), the heat kernel becomes a stationary
profile, and solutions that converge to a self-similar profile become localized with rapidly decaying tails in $\xi$. This suggests that learning in self-similar variables offers a clear spatial advantage: a fixed sampling window can capture the main part of the solution at large times.

The goal of this paper is to develop a precise self-similar-variable (SSV) formulation for heat-type equations that is compatible with neural-operator training, and to quantify the practical gains of learning in self-similar coordinates rather than in physical coordinates. To demonstrate that the approach applies across different heat-based PDEs, we specialize the SSV formulation to the two-dimensional incompressible Navier–Stokes equations and the one-dimensional viscous Burgers equation. To further show that the effect of the coordinate choice is architecture-independent, we present results using two simple fully connected models: a standard coordinate multi-layer perceptron (MLP) and a factorized fully connected network (FCN). 

Our comparisons across architectures show a consistent picture: within the
SSV framework, both the plain MLP and the factorized FCN produce significantly more accurate and stable extrapolation than their counterparts trained in physical coordinates. Moreover, we observe that SSV-trained models more reliably reproduce qualitative features that are intrinsic to the underlying dynamics--such as long-time physical patterns and, when present, coarsening/merging behaviors--whereas these effects are often degraded or missed by models trained directly in physical coordinates. 
This indicates that the main gain comes not from a sophisticated network design but from working in a coordinate system that is intrinsically adapted to the parabolic scaling and self-similar dynamics of the underlying equations. In this sense, our results provide evidence for the potential advantage of SSV-based training and point to a physically motivated strategy for learning solutions of other systems with self-similar asymptotics. 

Code to reproduce the experiments is available at \url{https://github.com/Kagami-319/SSV-Learning}.

\section{Networks}
In this section, we describe the two
network architectures used throughout the study. As a baseline we use MLP, which takes the concatenated coordinate
$z\in\mathbb{R}^m$ (either $(x,t)$ or $(\xi,\tau)$) and applies a single fully
connected network $f_\theta:\mathbb{R}^m\to\mathbb{R}$,
\[
  \mathcal{M}_\theta(z) = f_\theta(z).
\]
Our second model FCN is inspired by
branch--trunk constructions. Let
\[
  B_\theta : \mathbb{R} \to \mathbb{R}^K, \qquad
  T_\theta : \mathbb{R}^m \to \mathbb{R}^K
\]
denote two multilayer perceptrons (the branch and trunk subnetworks).
For a time-like coordinate $s\in\mathbb{R}$ (either $t$ or $\tau$) and a
space--time coordinate $z\in\mathbb{R}^m$ (either $(x,t)$ or $(\xi,\tau)$),
the FCN output is defined by
\[
  \mathcal{N}_\theta(s,z)
    = b_\theta + \sum_{k=1}^K B_{\theta,k}(s)\,T_{\theta,k}(z),
\]
where $b_\theta\in\mathbb{R}$ is a bias and $B_{\theta,k},T_{\theta,k}$ denote
the $k$-th components of $B_\theta$ and $T_\theta$.

\section{Learning the vorticity of the 2D Navier-Stokes equation}\label{3}

\subsection{System and self-similar variables}
In this section, we consider the 2D incompressible Navier-Stokes equations in vorticity form on $\R^2$ with viscosity normalized to $\nu=1$:
\begin{equation}\label{eq:NS-vort}
\partial_t \omega + u\cdot\nabla \omega = \Delta \omega, \qquad u = K*\omega,\qquad K(x) = \frac{1}{2\pi}\,\frac{x^\perp}{\abs{x}^2},\quad x^\perp=(-x_2,x_1).
\end{equation}
The circulation $\Gamma:=\int_{\R^2}\omega(x,t)\,\dd x$ is conserved. Noting that the scaling
$\omega_\lambda(x,t)=\lambda^2\,\omega(\lambda x,\lambda^2 t)$, $\, u_\lambda(x,t)=\lambda\,u(\lambda x,\lambda^2 t)$ leaves \eqref{eq:NS-vort} invariant, we are inspired to introduce self-similar variables
\begin{equation*}\label{eq:ssv}
\xi=\frac{x}{\sqrt {t+1}},\quad \tau=\log (t+1),\quad \omega(x,t)=(t+1)^{-1}\,\Omega(\xi,\tau),\quad u(x,t)=(t+1)^{-1/2}\,U(\xi,\tau).
\end{equation*}
A direct change of variable yields
\begin{equation}\label{eq:ssv-eq}
\boxed{\, \partial_\tau\Omega + U\cdot\nabla_\xi\Omega = \cL\Omega,\qquad U=K*\Omega,\, }\qquad \cL:=\Delta_\xi + \tfrac12\,\xi\cdot\nabla_\xi + \mathrm{Id}.
\end{equation}
The Oseen vortex $\displaystyle G(\xi)=\frac{1}{4\pi}\e^{-\abs{\xi}^2/4}$ satisfies $\cL G=0$; moreover $U^G\cdot\nabla G\equiv0$ by orthogonality of the tangential $U^G$ and radial $\nabla G$, hence $\Omega\equiv \alpha G$ is a steady solution of \eqref{eq:ssv-eq} for any circulation $\alpha$.

\subsection{Training in the self-similar framework}
It was shown in \cite{gallay_wayne_2002} that, starting from a $L^1$ initial data, the vorticity will converge to the Gaussian at large time:
\begin{proposition}[Long-time Behavior]\label{prop1}
    Fix $\mu\in(0,\tfrac{1}{2})$. There exist positive constants $r_2$ and $C$ such that, for any initial data $\Omega_0$ with $\|\Omega_0\|_2\leq r_2$, the solution $\Omega(\cdot,\tau)$ of \eqref{eq:ssv-eq} satisfies:
    \begin{align}
        \left\|\Omega(\cdot,\tau)-\left(\int_{\R^2}\Omega(\xi)d\xi\right) G(\cdot)\right\|_2\leq Ce^{-\mu\tau},\quad \tau\ge0.
    \end{align}
\end{proposition}

Thus, by Proposition \ref{prop1}, we shall see
\[
    |\Omega(\xi,\tau)| \lesssim e^{-c|\xi|^2}
    \quad\text{for some }c>0,
\]
uniformly for large $\tau$.
Therefore the self–similar solution is localized in $\xi$, and the error
outside a moderate self–similar radius decays exponentially.
This motivates measuring the approximation error on a fixed self–similar window
and ignoring the exponentially small tails. Fix $C>1$ and denote the 2D window
\[
  D_C := \{\xi\in\R^2:\ |\xi|\le C\}.
\]
For the error field $\mathcal{E}_{\mathrm{ssv}}(\xi,\tau)=\Omega(\xi,\tau)-\widehat\Omega(\xi,\tau)$
we define the windowed $L^2$ error as
\begin{equation}\label{eq:A}
  E_{ssv}(C;\tau) \ :=\ \int_{D_C} |\mathcal{E}_{\mathrm{ssv}}(\xi,\tau)|^2\dd\xi.
\end{equation}

\subsection{Physical-space window (expanding disk)}\label{sec:phys-window}

In physical coordinates the vorticity field $\omega(x,t)$ spreads diffusively:
for the linear heat equation the fundamental solution is the 2D Gaussian
\[
    G_t(x) = \frac{1}{4\pi t} \exp\!\Bigl(-\frac{|x|^2}{4t}\Bigr),
\]
whose effective support has radius of order $\sqrt{t}$. For general 2D incompressible Navier-Stokes
vorticity, the mild formulation
\[
    \omega(t) = e^{t\Delta}\omega_0
    - \int_0^t e^{(t-s)\Delta}\bigl(u\cdot\nabla\omega\bigr)(s)\,ds
\]
shows that each Duhamel term is obtained by convolving data with the same heat
kernel. Thus, it is natural to regard the core of
$\omega(\cdot,t)$ as living on a disk of radius comparable to $\sqrt{t+1}$, and
the contribution of the far field as exponentially small in $|x|^2/(t+1)$. This picture is perfectly consistent with the self–similar formulation where we restrict the sampling area inside a fixed disk.

To compare self–similar and physical training in a fair way, we choose the
physical window so that it is exactly the image of $D_C$ under the
self–similar change of variables. For any $t\ge 0$ and $C>0$ we define
\begin{equation*}\label{eq:DtC-def}
    D_{t,C} := \{x\in\mathbb{R}^2 : |x|\le C\sqrt{t+1}\},
\end{equation*}
which is the expanding disk obtained from $D_C$ by the map
\begin{equation}\label{eq:x-xi-map}
    x = \sqrt{t+1}\,\xi.
\end{equation}
If $\xi$ is sampled uniformly from $D_C$, then $x$ is sampled uniformly from
$D_{t,C}$; the Jacobian of \eqref{eq:x-xi-map} is a constant factor
$(t+1)$ and does not depend on the direction of $\xi$. Consequently, the
same batch of self–similar samples $\{\xi_j\}_{j=1}^M$ used to train the SSV
model induces, via \eqref{eq:x-xi-map}, a batch of physical points
$\{x_j\}_{j=1}^M$ that are uniformly distributed on $D_{t,C}$.

For the physical error field $\mathcal{E}_{\mathrm{phy}}(x,t)
    := \omega(x,t) - \widehat{\omega}(x,t)$,
we therefore measure the continuous windowed $L^2$ error by
\begin{equation}\label{eq:E-phys-Ct}
    E_{phy}(C;t) := \int_{D_{t,C}}
              \bigl|\mathcal{E}_{\mathrm{phy}}(x,t)\bigr|^2\,dx.
\end{equation}

\subsection{Fair comparison protocol}\label{subsec:fair-comparison}
To compare SSV and physical training in a coordinate--agnostic way we adopt
the following protocol. 

\begin{enumerate}
  \item \textbf{Architecture parity.}
  The SSV and physical models share the same network architecture. For each architecture, we train two independent instances—one in self-similar coordinates and one in physical coordinates—while keeping all hyperparameters identical, including depth, width, optimizer, learning-rate schedule, training time window, batch size, and the total number of training steps.

  \item \textbf{Sampling (one–to–one between coordinates).}
  We draw i.i.d paired samples $\{(\xi_i,\tau_i)\}_{i=1}^M$ by Monte Carlo sampling on the training window
\[
\tau_i \sim {\rm Unif}\big[\log(1+t_{\min}),\log(1+t_{\max})\big],\qquad
\xi_i \sim {\rm Unif}(D_C),
\]
and then set the corresponding physical variables by $t_i = e^{\tau_i} - 1$ and $x_i = \sqrt{t_i+1}\,\xi_i$.
The linear scaling $\xi \mapsto x$ sends the uniform distribution on $D_C$ to the uniform
distribution on $D_{t,C}$. Hence, for each sampled pair $(\xi_i,\tau_i)$ used by the SSV model, there is a
one-to-one matched physical sample $(x_i,t_i)$ seen by the physical model.

\item \textbf{Loss functions.}
Our target is to minimize the \emph{time-averaged windowed} mean-squared errors
\[
L_{\mathrm{ssv}}(C)
:= \mathbb{E}_{\tau}\Big[E_{\mathrm{ssv}}(C;\tau)\Big],
\qquad
L_{\mathrm{phy}}(C)
:= \mathbb{E}_{\tau}\Big[E_{\mathrm{phy}}(C;t)\Big].
\]

In practice, we approximate the above expectations by
\[
\widehat L_{\rm ssv}
:=\frac{1}{M}\sum_{i=1}^M\bigl|\widehat \Omega(\xi_i,\tau_i)-\Omega(\xi_i,\tau_i)\bigr|^2,
\qquad
\widehat L_{\rm phy}
:=\frac{1}{M}\sum_{i=1}^M\bigl|\widehat\omega(x_i,t_i)-\omega(x_i,t_i)\bigr|^2.
\]

  \item \textbf{Extrapolation metric (relative MSE).}
  During extrapolation, we quantify predictive accuracy using the
  relative mean squared error (MSE) evaluated in the physical
  domain $D_{t,C}$. Concretely, for a predicted field $\widehat{\omega}$
  and the reference solution $\omega$ at time $t$, we report
  \begin{equation}\label{eq:rel-mse-def}
    \mathrm{RelMSE}(t)
    :=
    \frac{\mathbb{E}\bigl[(\widehat{\omega}(t,x)-\omega(t,x))^2\bigr]}
         {\mathbb{E}\bigl[\omega(t,x)^2\bigr]},
  \end{equation}
  where the expectation is taken with respect to the uniform distribution on
  $D_{t,C}$ (approximated by averaging over the evaluation grid points).
\end{enumerate}

\subsection{Experiments}
Our experiments are designed to evaluate the practical advantage of training in SSV for extrapolation. We first compare SSV-based and physical-coordinate training under a fixed protocol, and then replicate the same study with two different network architectures to show that any observed gain is model-independent. 

We use the same initial condition throughout, given by a superposition of
two Gaussians. More precisely, we set 
\[
  \omega_0(x,y)
  = A_1 \exp\!\left(-\frac{(x-x_1)^2 + (y-y_1)^2}{\sigma_1^2}\right)
  + A_2 \exp\!\left(-\frac{(x-x_2)^2 + (y-y_2)^2}{\sigma_2^2}\right),
\]
with $A_1=1$, $A_2=0.6$, $(x_1,y_1)=(-1.5,0.5)$, $(x_2,y_2)=(1.0,-0.8)$,
$\sigma_1=1$, and $\sigma_2=1.3$, which yields an essentially non-radial
initial vorticity field.

The long-time behavior for such initial data is clear. By Proposition~\ref{prop1}, as $t$ increases,  the two Gaussian components move toward each other and eventually ``merge" into a single-Gaussian profile in the long run. After the solution enters this merged regime, learning in SSV becomes comparatively easy, since the target dynamics has largely stabilized in SSV and is close to a time-independent profile. For this reason, we choose the training window $t\in[0,0.3]$, terminating before the ``merging" completes. In this way, extrapolation genuinely tests whether the model has learned the structural feature of the solution, and therefore produces physically faithful long-time predictions, rather than simply fitting the near-steady merged regime.

\begin{figure}[ht]
  \centering
  \begin{subfigure}{0.48\textwidth}
    \centering
    \includegraphics[width=\linewidth]{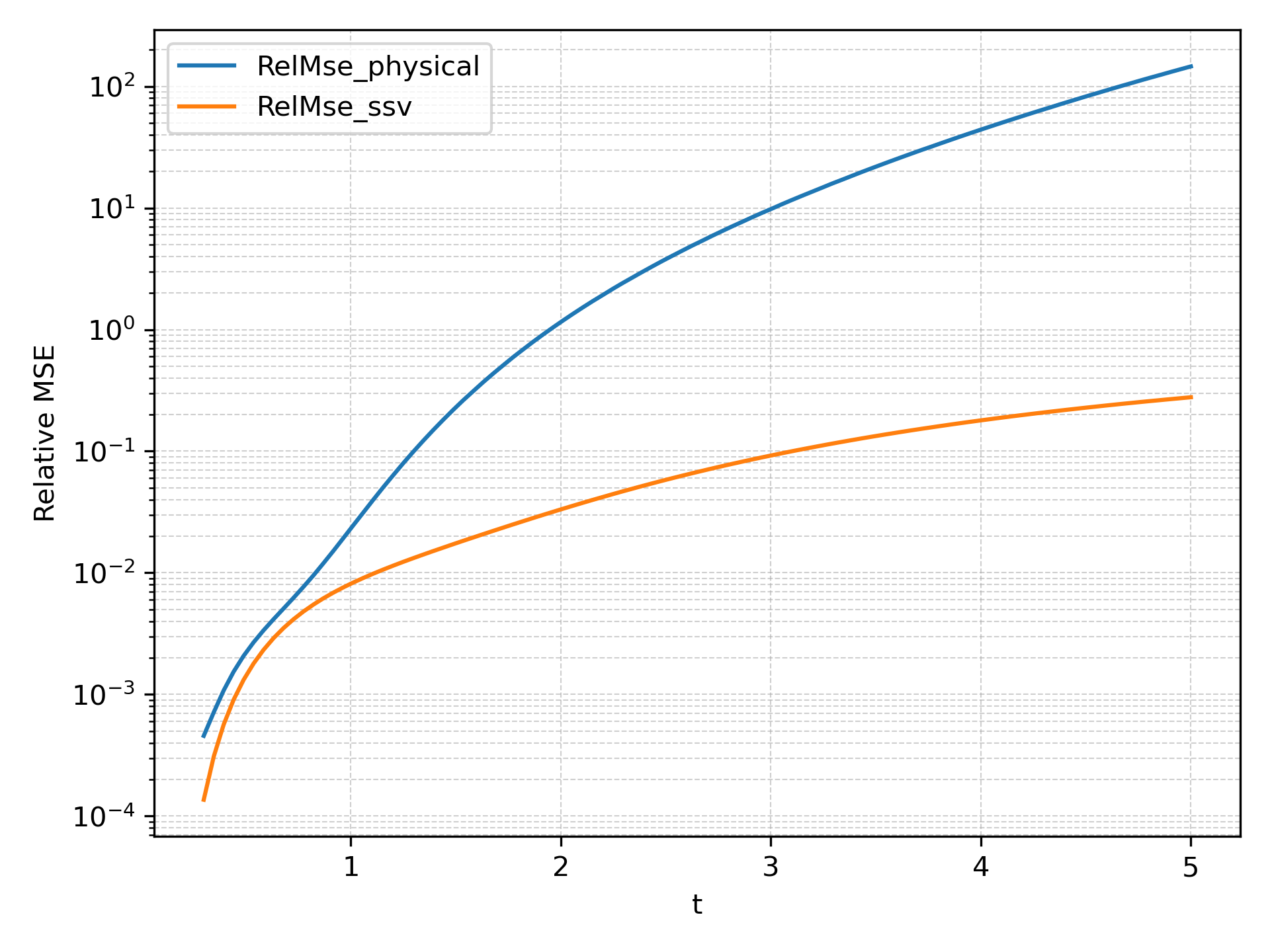}
    \caption{FCN}
    \label{fig:fcn_rmse-ns}
  \end{subfigure}\hfill
  \begin{subfigure}{0.48\textwidth}
    \centering
    \includegraphics[width=\linewidth]{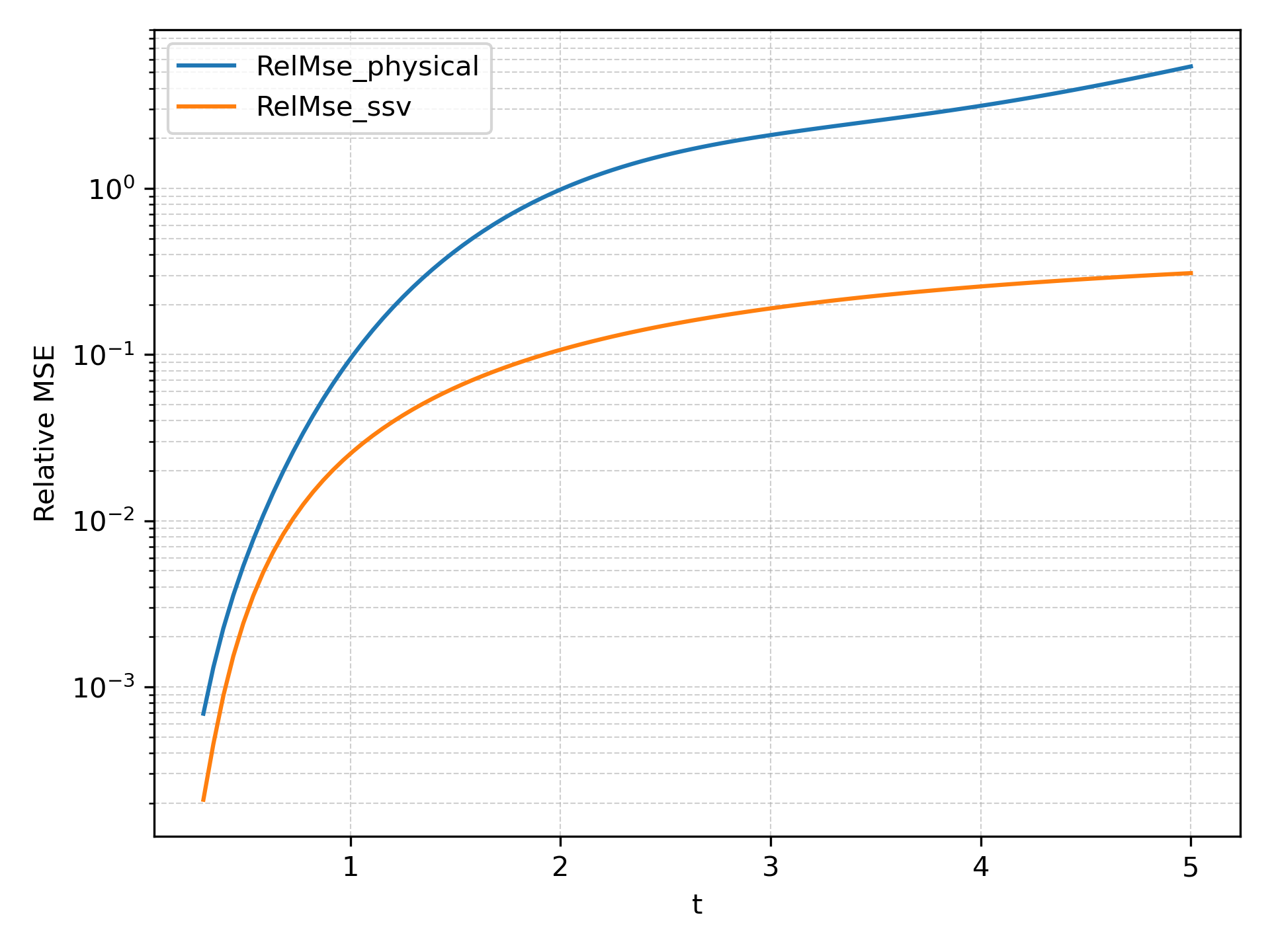}
    \caption{MLP}
    \label{fig:mlp_rmse-ns}
  \end{subfigure}
  \caption{Relative MSE over $t\in[0.3,5]$ under two architectures.}
  \label{fig:rmse_compare-ns}
\end{figure}

\begin{figure}[ht]
  \centering
  \includegraphics[width=0.9\textwidth]{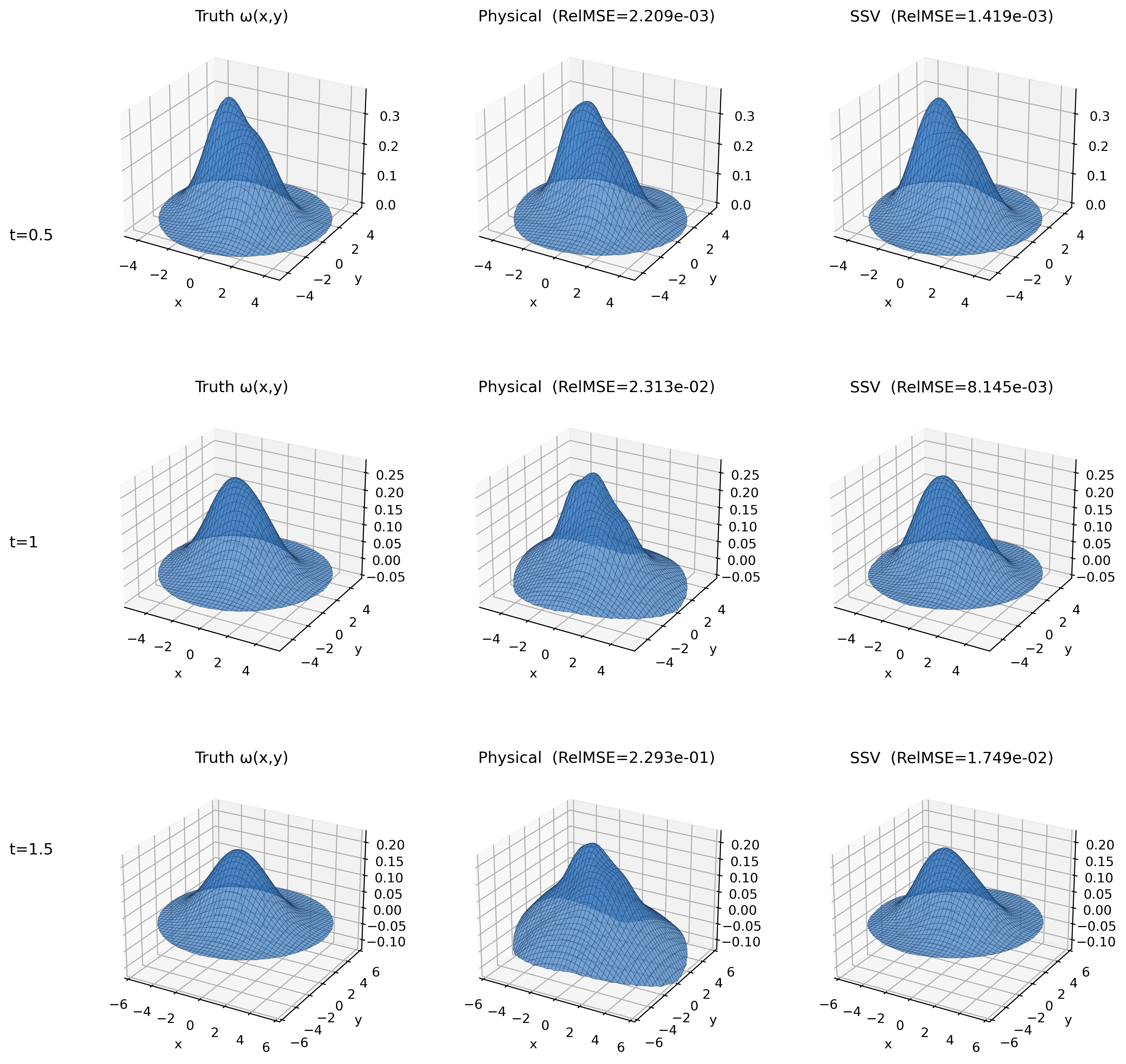}
  \caption{Example of extrapolation to $t=0.5,1,1.5$ under the FCN:
  truth (left), physical head (middle), and self-similar head mapped to physical
  space (right).}
  \label{fig:fcn-ns}
\end{figure}

\begin{figure}[ht]
  \centering
  \includegraphics[width=0.9\textwidth]{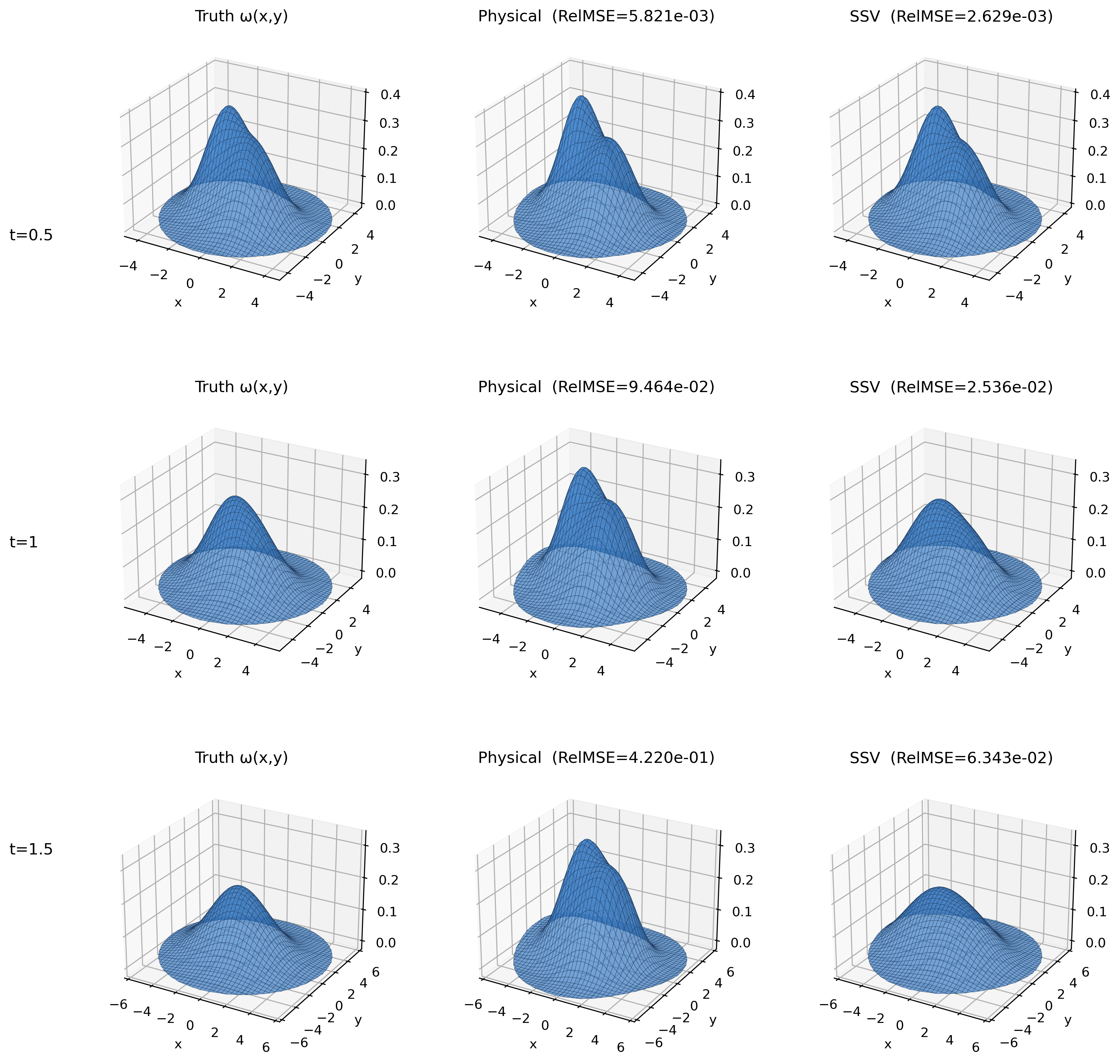}
  \caption{Example of extrapolation to $t=0.5,1,1.5$ under the MLP.}
  \label{fig:mlp-ns}
\end{figure}

Figure~\ref{fig:rmse_compare-ns} shows the relative MSE for $t\in[0.3,5]$ using FCN and MLP respectively; Figure~\ref{fig:fcn-ns} \& ~\ref{fig:mlp-ns} represent some chosen examples during the extrapolation ($t=0.5,1,1.5$) for the FCN and MLP respectively, where each row corresponds to a fixed time out of the training interval, and the columns display the reference vorticity $\omega(t,x,y)$, the prediction of the physical head, and the prediction of the self-similar head (which has been mapped back to physical variables for visual comparison). 

We shall see that, for both architectures and across both short- and long-horizon extrapolation, the self-similar head consistently delivers markedly more accurate predictions than the physical head. More importantly, the SSV-trained models clearly capture the physically relevant "merging" trend. In contrast, the physical head fails to reproduce this qualitative transition, instead developing pronounced artifacts whose discrepancies grow more evident as $t$ increases. Together, these observations support our expectation that learning in the self-similar framework better captures the structure of the solution and reflects the long-time dynamics of the underlying system. 

\section{Viscous Burgers}\label{4}

In this section, we check the advantages of the SSV framework for the one-dimensional viscous Burgers equation which is as well heat--based, but has a much simpler nonlinearity.

\subsection{Equation Setup}

We work on the whole line $x \in \mathbb{R}$ with viscosity normalized to~$1$ and consider
the standard viscous Burgers equation
\begin{equation}\label{eq:burgers}
  \partial_t u + \frac{1}{2}\,\partial_x\!\bigl(u^2\bigr)
  \;=\; \partial_{xx} u,
  \qquad x \in \mathbb{R},\ t > 0.
\end{equation}

To align this example with the SSV framework of Section \ref{3}, we use the same time change
\begin{equation}\label{eq:ssv-burgers-change}
  \xi = \frac{x}{\sqrt{t+1}}, \qquad
  \tau = \log(t+1),
\end{equation}
and introduce the self--similar unknown $\omega$ by
\begin{equation*}\label{eq:ssv-burgers-scaling}
  u(x,t) = (t+1)^{-1/2} \omega(\xi,\tau).
\end{equation*}
A direct computation shows that $\omega$ solves the SSV Burgers equation
\begin{equation}\label{eq:ssv-burgers-pde}
  \partial_\tau \omega + \omega \,\partial_\xi \omega
  \;=\; \partial_{\xi\xi} \omega + \frac{1}{2}\,\xi\,\partial_\xi \omega
            + \frac{1}{2}\, \omega.
\end{equation}
It is well known that, for integrable initial data with nonzero mass, the long-time asymptotics of viscous Burgers are described by the diffusion wave. 
\begin{proposition}[Long-time Behavior; \cite{kim_tzavaras_2001}]\label{prop2}
    Let $\omega$ be the solution to the Cauchy problem \eqref{eq:ssv-burgers-pde} with initial data $\omega_0\in L^1\cap L^\infty$, $\partial_\xi \omega_0\in L^2$, and total mass
    $\displaystyle \int_{\mathbb R} \omega_0(\xi)\,d\xi = M_0$.
Then
\begin{equation}\label{eq:3.5}
\omega(\xi,\tau)\to \mathcal{G}_{M_0}(\xi)\qquad \text{as } \ \ \ \tau\to\infty,
\end{equation}
a.e.\ and in $L^1(\mathbb R)$, where $\mathcal{G}_{M_0}$ is the diffusion wave given by 
\begin{equation}\label{diffusion wave}
\mathcal{G}_M(\xi)
=
\frac{\,\bigl(1-e^{-M/2}\bigr)\,e^{-\xi^{2}/4}}
{\,1-\bigl(1-e^{-M/2}\bigr)\,\dfrac{1}{\sqrt{\pi}}
\displaystyle\int_{-\infty}^{\xi/2} e^{-\zeta^{2}}\,d\zeta\,},
\qquad M\in\mathbb{R}.
\end{equation}
\end{proposition}

Our experiments therefore probe how well the networks capture this non-Gaussian self-similar profile. 

\subsection{Sampling windows and training loss}

The SSV and physical sampling windows are one-dimensional analogues of the disks used in
Section~3.
For a fixed $C > 0$ we set
\begin{equation*}
  I_C := \{\xi \in \mathbb{R} : |\xi| \le C\},
  \qquad
  I_{t,C} := \{x \in \mathbb{R} : |x| \le C\sqrt{t+1}\}.
\end{equation*}

Let $u^\ast(x,t)$ denote the reference solution obtained from a numerical
solver, and let $u(x,t)$ and $\omega(\xi,\tau)$ be the
predictions of the physical and SSV models, respectively. For each fixed time $t$ we
measure the windowed $L^2$ errors
\begin{equation*}
    \begin{aligned}
     E_{\mathrm{phy}}(C;t) &:= \int_{I_{t,C}} \bigl|u^\ast(x,t) - u(x,t)\bigr|^2 \, dx,\\
  E_{\mathrm{ssv}}(C;\tau) &:= \int_{I_C} \bigl| e^{\tau/2} u^\ast(e^{\tau/2}\xi ,e^{\tau}-1) - \omega(\xi,\tau)\bigr|^2 \, d\xi.
    \end{aligned}
\end{equation*}

In practice, we form a Monte Carlo training set by drawing i.i.d.\ paired samples $\{(t,x)\}_{j=1}^N$.
Specifically, we sample physical time $t_j$ uniformly on $[t_{\min},t_{\max}]$ and
sample spatial locations $x_j$ by drawing indices uniformly from the reference spatial grid. For the SSV model, each sampled pair $(x_j,t_j)$ is mapped to the corresponding self-similar coordinates. Similar to the Navier-Stokes case, our target is to minimize the time-averaged windowed mean-squared errors
\[
L_{\rm ssv}(C):=\mathbf{E}_{t}\!\big[E_{\rm ssv}(C;\tau)\big],\qquad
L_{\rm phy}(C):=\mathbf{E}_{t}\!\big[E_{\rm phy}(C;t)\big],
\]
which are approximated by
\[
\widehat L_{\rm ssv}
:=\frac{1}{N}\sum_{j=1}^N
\bigl|\widehat \omega(\xi_j,\tau_j)-\omega(\xi_j,\tau_j)\bigr|^2,
\qquad
\widehat L_{\rm phy}
:=\frac{1}{N}\sum_{j=1}^N
\bigl|\widehat u(x_j,t_j)-u(x_j,t_j)\bigr|^2.
\]

\subsection{Experiments}
\begin{figure}[ht]
  \centering

  \begin{subfigure}{0.5\textwidth}
    \centering
    \includegraphics[width=\linewidth]{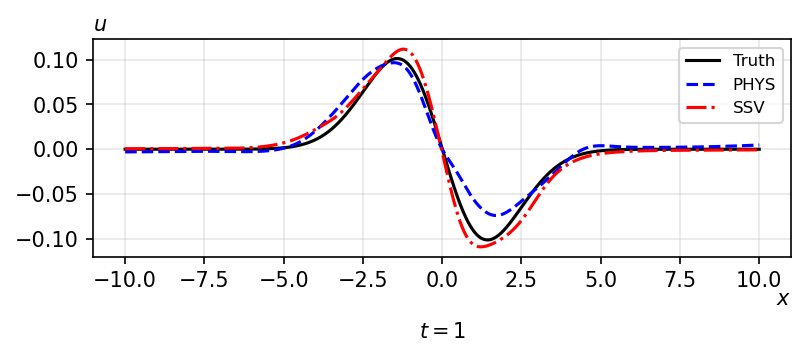}
  \end{subfigure}\hfill
  \begin{subfigure}{0.5\textwidth}
    \centering
    \includegraphics[width=\linewidth]{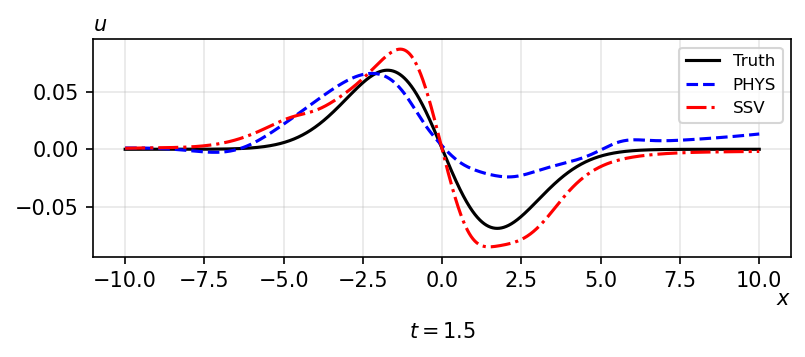}
  \end{subfigure}

  \vspace{0.4em}

  \begin{subfigure}{0.5\textwidth}
    \centering
    \includegraphics[width=\linewidth]{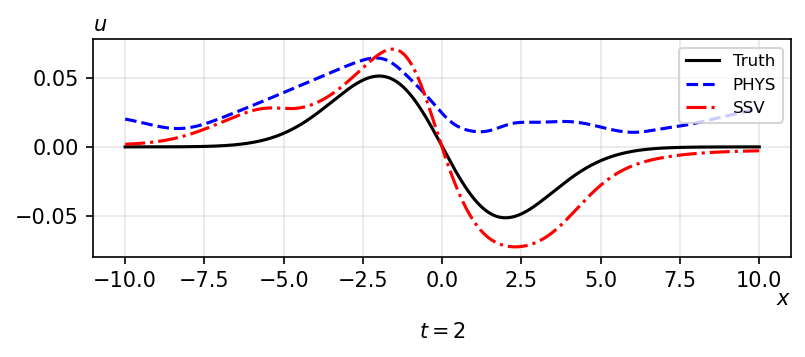}
  \end{subfigure}\hfill
  \begin{subfigure}{0.5\textwidth}
    \centering
    \includegraphics[width=\linewidth]{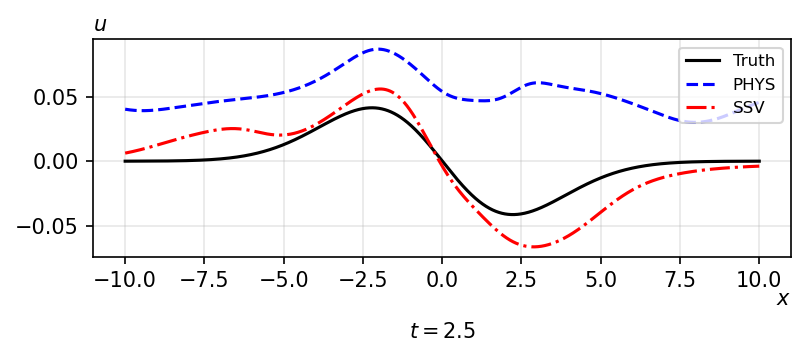}
  \end{subfigure}

\caption{Comparison between the physical FCN
           and the SSV FCN at t=1, 1.5, 2, 2.5.}
  \label{fig:burgers-fcnet-1d}
\end{figure}

\begin{figure}[ht]
  \centering

  \begin{subfigure}{0.5\textwidth}
    \centering
    \includegraphics[width=\linewidth]{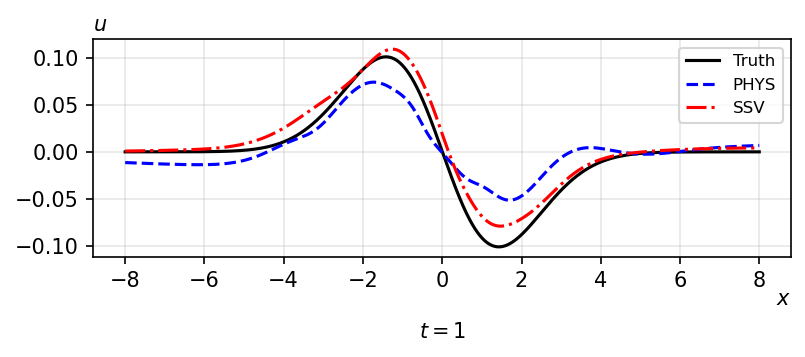}
  \end{subfigure}\hfill
  \begin{subfigure}{0.5\textwidth}
    \centering
    \includegraphics[width=\linewidth]{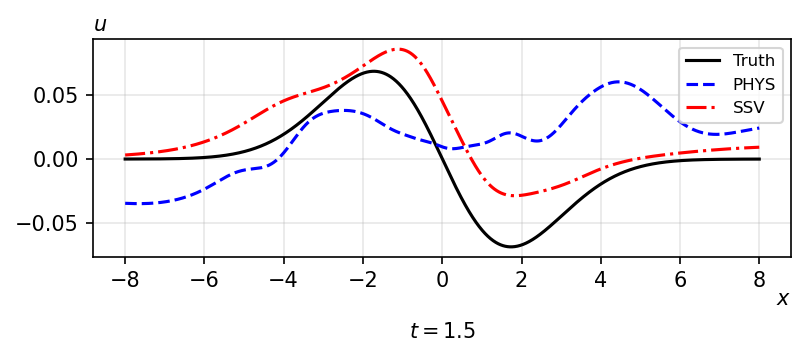}
  \end{subfigure}

  \vspace{0.4em}

  \begin{subfigure}{0.5\textwidth}
    \centering
    \includegraphics[width=\linewidth]{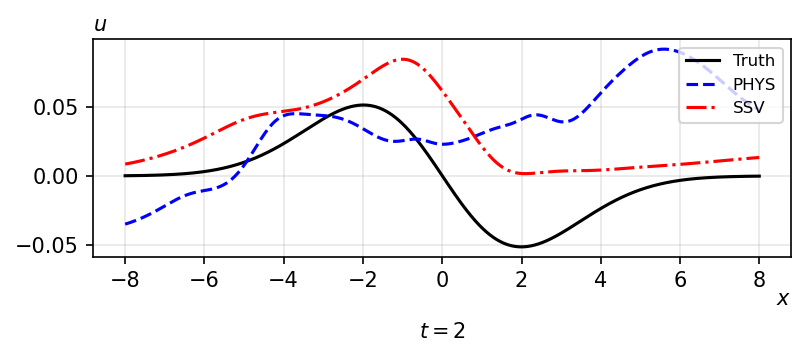}
  \end{subfigure}\hfill
  \begin{subfigure}{0.5\textwidth}
    \centering
    \includegraphics[width=\linewidth]{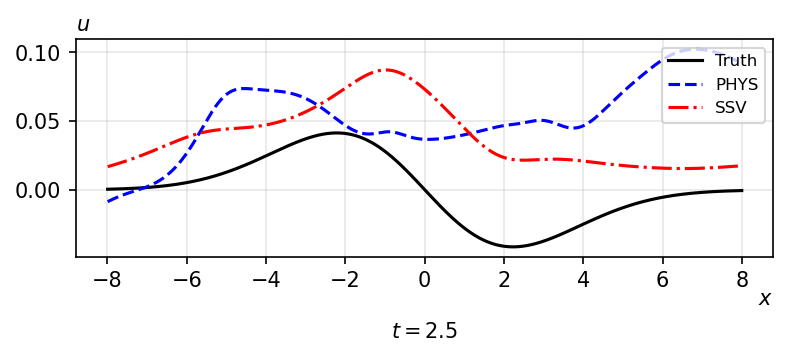}
  \end{subfigure}

  \caption{Same experiment as in Figure~\ref{fig:burgers-fcnet-1d} for MLP.}
  \label{fig:burgers-concat-1d}
\end{figure}

The initial data is chosen to be a compactly supported bipolar box profile,
\[
u(x,0) = u_0(x) =
\begin{cases}
  1,  & -1 < x < 0,\\[2pt]
 -1, & 0 < x < 1,\\[2pt]
  0,  & |x| \ge 1 .
\end{cases}
\]
Both models are trained on the time window $t\in[0,0.5]$. We deliberately keep this window short as we did in Section~3, so that in SSV the target remains genuinely time-dependent.
Figure~\ref{fig:rmse_compare-ns} reports the relative MSE over $t\in[0.5,5]$. Figures~\ref{fig:burgers-fcnet-1d} and \ref{fig:burgers-concat-1d} further visualize the corresponding extrapolation behavior. 
In Figures~\ref{fig:burgers-fcnet-1d} and \ref{fig:burgers-concat-1d}, each panel corresponds to an evaluation time $t\in \{ 1,1.5,2,2.5 \}$ beyond the training horizon. For each time, we plot the reference solution (truth) together with the predictions from the physical and SSV heads, using the same spatial interval for all curves.

\begin{figure}[ht]
  \centering
  \begin{subfigure}{0.48\textwidth}
    \centering
    \includegraphics[width=\linewidth]{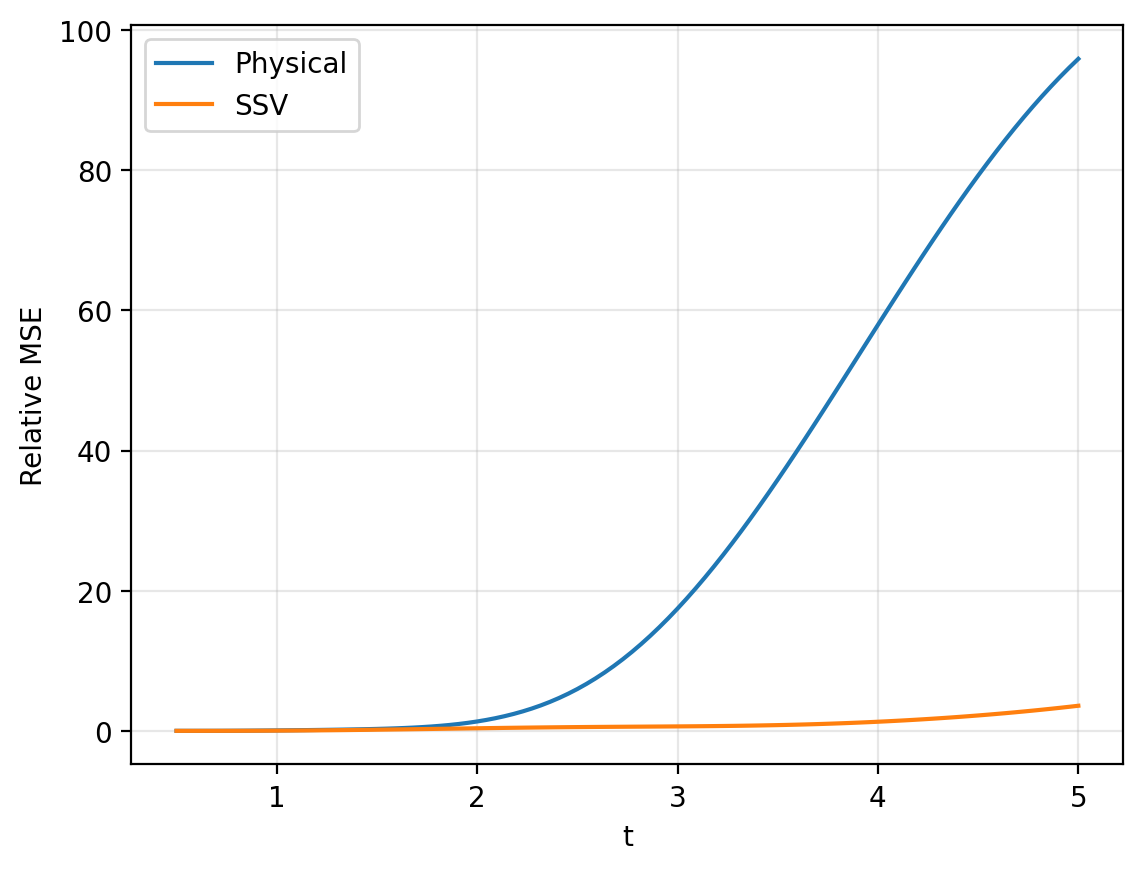}
    \caption{FCN}
    \label{fig:fcn_rmse-burgers}
  \end{subfigure}\hfill
  \begin{subfigure}{0.48\textwidth}
    \centering
    \includegraphics[width=\linewidth]{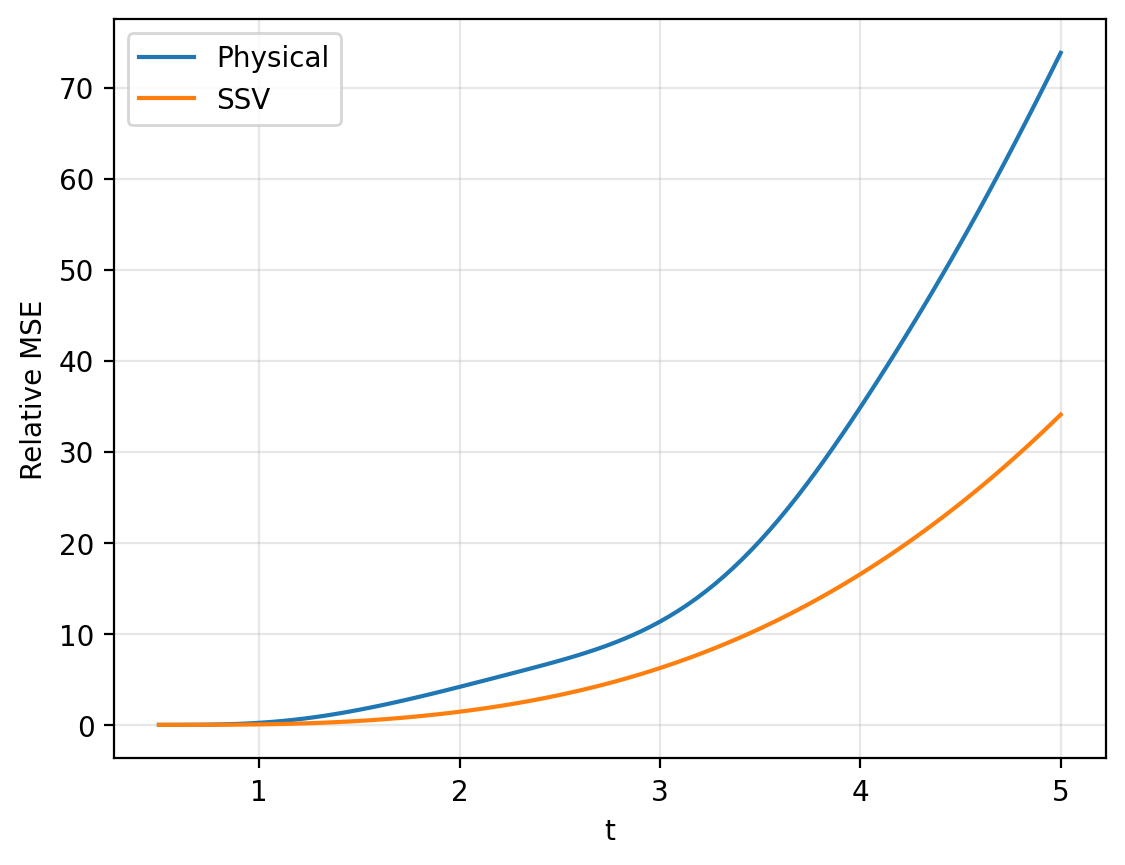}
    \caption{MLP}
    \label{fig:mlp_rmse-burgers}
  \end{subfigure}
  \caption{Relative MSE over $t\in[0.5,5]$ under two architectures.}
  \label{fig:rmse_compare-burgers}
\end{figure}

We observe a consistent qualitative pattern, fully analogous to the two-dimensional Navier-Stokes experiments: The SSV framework preserves the shape of the viscous shock profile well beyond the training horizon while the physical framework shows increasingly pronounced distortions as $t$ grows. In other words, when we fix the architecture and training budget, the SSV representation again yields significantly more reliable long-time extrapolation. This one-dimensional
example confirms that the advantage of learning in self--similar variables is not an artifact of the Navier-Stokes system, but rather a robust feature of heat-based evolution equations.

\section{Conclusion}\label{sec:conclusion}

The results in Sections \ref{3} and \ref{4} consistently indicate a robust advantage of SSV training. Models trained in self-similar coordinates not only extrapolate more accurately and stably, but also more reliably reproduce qualitative long-time features that are intrinsic to the underlying dynamics. Importantly, this improvement does not primarily come from architectural sophistication; rather, it is driven by an appropriate representation that aligns with the self-similar structure of heat-based evolution. In this sense, self-similar coordinates act as a mathematically motivated inductive bias.

More broadly, our findings suggest a general guideline for PDE operator learning: taking advantage of the physical intuition of the system can be more effective than further architectural engineering. From this perspective, SSV-based training provides a practical route to improved rollout stability and long-time prediction for heat-based systems, and the same strategy extends to other equations with self-similar asymptotics such as the porous medium equation $\partial_t u=\Delta u^m$,$m\in\N$, which admits self-similar spreading solutions (see  \cite{Barenblatt1996}).

Looking ahead, several directions appear natural: extending the framework to broader classes of systems with known scaling laws; developing adaptive windowing strategies for settings where asymptotic localization is weaker; combining SSV-based training with physics-based constraints to further improve robustness; and establishing theory that connects SSV-based sampling and coordinate choices to generalization and rollout stability in operator learning.

\newpage
\bibliography{iclr2026_conference.bib}

\end{document}